# Building of Networks of Natural Hierarchies of Terms Based on Analysis of Texts Corpora

*D.V. Lande, Institute of Data Recording Problems, NAS Ukraine*

The technique of building of networks of hierarchies of terms based on the analysis of chosen text corpora is offered. The technique is based on the methodology of horizontal visibility graphs. Constructed and investigated language network, formed on the basis of electronic preprints arXiv on topics of information retrieval.

Ontologies creation on specific expertise areas is currently a relevant task. Obviously, large industrial ontology construction is a complex problem that requires large resource expenses. In any case, a certain stage of common ontologies construction is the relevant thesauruses and terminological ontologies construction.

The method of building a network of natural terms hierarchy is proposed which may be regarded as "quasiontology", i.e. the basis for corresponding terminological ontology formation. Natural terms hierarchy network of is based on «significantly informative» text elements, the reference words and phrases. The methodology to identify such terms is given in [1, 2]. The use of such elements can form search images and cover the whole knowledge bases for the further common ontology construction. Reference words and phrases for natural terms hierarchy construction are selected with taking into account the discriminant power. However, one of the properties is not sufficient for the construction of thesauruses and ontologies. Sometimes words with low discriminant power, in particular, the most frequent words of the given subject area (e.g., "Information", "Retrieval", "Search" words in the information retrieval body) are essential for a task that is considered.

Network of natural hierarchies of terms (NNHT) creation is based on the text corpuses content of the appropriate orientation. In this case "naturality" of terms hierarchy is understood as an abandonment of special methods of semantic analysis while network formation. All links in such network are determined by the natural use of the words and phrases that are extracted from the text corpuses of statistically significant volumes. Network of natural hierarchies of terms created automatically can be considered as a basis for further automated terminological ontologies generation.

Network of natural hierarchies of terms generation algorithm considers iterations implementation, covering pre-processing of source text corpus, terms definition and sorting, most tangible terms required number selection (the largest nodes compacted horizontal visibility graph [3]), NNHT construction and representation. Let's consider these steps in detail.

1. First we need to select source text corpus. The example of such case is electronic preprints annotation array for arXiv (www.arxiv.org) for the period of 2007-2010 years, selected by category of information retrieval (cap. cs.IR) of 550 records. Such text corpus preprocessing provides the text parts of records selection, non-text characters exception and stemming.

2. The second stage includes association of "discriminant power" (TFIDF) to every single word of the text corpus, which equals to the canonical multiplication of the word frequency in the text fragment (Term Frequency) to the binary logarithm of the reciprocal of the text fragments number in which this word is encountered (Inverse Document Frequency) [4].

3-4. Performed the same way as in the previous step for two-word phrases (bigrams) and three words (trigrams).

5. Compacted horizontal visibility graphs (CHVG) are built for terms sequences and their weight values based on TFIDF [1, 2], and then the words weight values re-definition is done. This procedure allows to take into account not only significant discrimination power terms, but also high frequency terms, which are important to the overall text corpus theme. Then all the terms are sorted descending calculated weight values of CHVG corresponding nodes.

The terms of the so-called stop-dictionary is not a subject of further analysis. This is usually a fixed set of words, which do not play a significant role in the text content.

6. The required volume NNHT (number N) is expertly determined, by then a corresponding number of single words, bigrams and trigrams (of N + N + N elements) with the highest weight values CHVG is selected.

7. The network of natural hierarchies of terms are built based on elements obtained on the previous step, in which nodes are considered as the terms themselves and links correspond to the occurrence of certain terms to the other ones. Fig. 1 illustrates the principle of building NNHT-links. Individual geometric shapes in this illustration correspond to single words.

First row corresponds to a selected set of single words, the second - the set of bigrams, and the third - the set of trigrams. If a single word is included to the bigram or trigram, or bigram is included to trigram, the link is formed denoted by the arrow. A nodes set, which corresponds to the terms, and links form a three-tier NNHT.

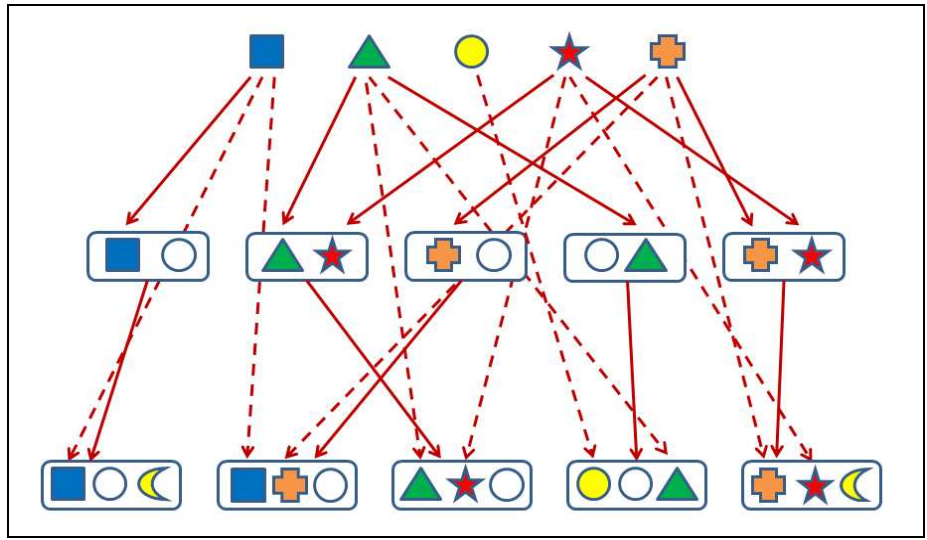

Fig. 1 – Links building in three-tier NNHT

8. At the last stage of NNHT creation, it is mapped to analytical and complex networks visualization software. The incidence matrix of conventional csv format is created for loading natural terms hierarchy networks to the databases.

Outgoing degrees of nodes distribution was determined for obtained natural terms hierarchy networks of different sizes depending on selected text corpus, which appeared to be close to exponential ($p(k)= Ck^{\alpha}$), i.e. such networks are scale-free. It turned out, that α coefficient for networks of different sizes (from 20 +20 +20 to 200 +200 + 200) is from 2.1 to 2.3. Fig. 2. represents a small NNHT with size 20+20+20, which is visualized as a spiral by author's suggested method.

Fig. 3 is a perspective view of a network of 200+ 200+ 200, which is visualized by Gephi system (https://gephi.org/).

Fig. 4 shows fragments of the that match the basic terms.

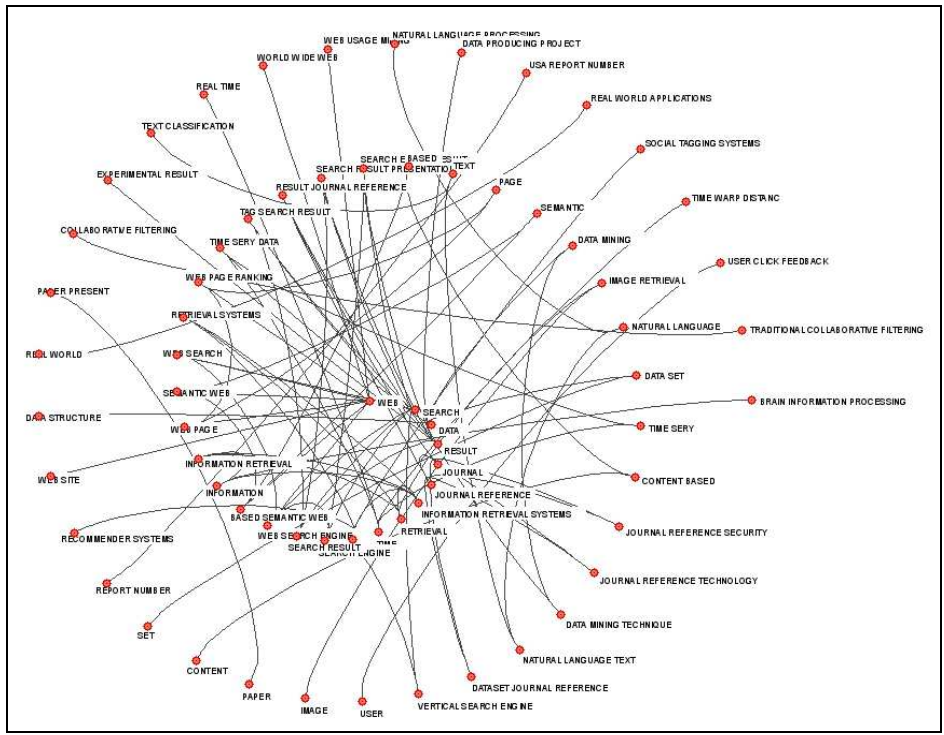

Fig. 2 – NNHT of size 20+20+20

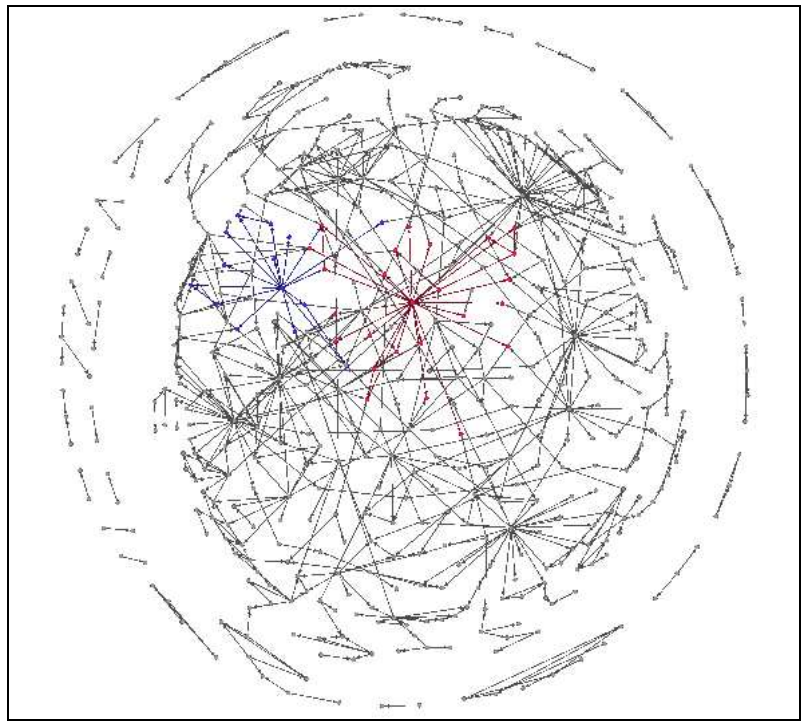

Fig. 3 – NNHT of 200+ 200+ 200 visualization by Gephi system

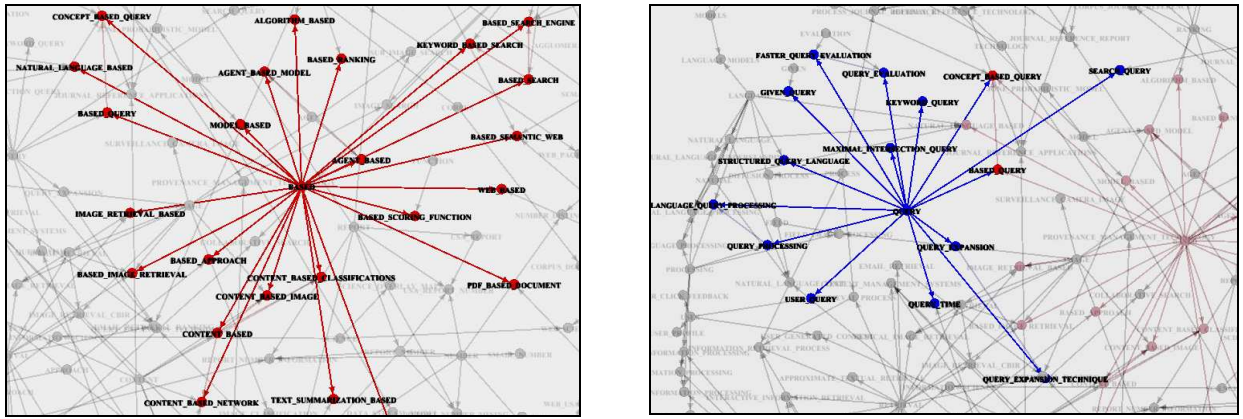

Fig. 4 – NNHT fragments

Thus, the results of the research are:
- Network of natural hierarchies of terms creation algorithm proposed based text corpuses analysis.
- Network of natural hierarchies of terms is built with proposed algorithm.
- Network of natural hierarchies of terms properties are investigated. The networks appeared to be scale-free by outgoing relations.
- Network of natural hierarchies of terms renderers are selected.
- Language network constructed using the proposed technique can be used as a base for building a general ontology (in this example – on information retrieval subject), as a ready-to-use navigation tools in relevant topics databases, and for context prompts for users of information retrieval system.